\documentclass[journal]{IEEEtran}

\usepackage{booktabs,amsmath,algorithm,algorithmic,multirow,graphicx,epstopdf} 

\usepackage{amssymb} 
\usepackage{color}
\usepackage{subfigure}
\usepackage{cite}
\usepackage{graphicx}
\usepackage[colorlinks, linkcolor=red]{hyperref}

\newcommand{\ie}{\emph{i.e.~}}

\usepackage{subfigure} 

\usepackage{comment}

\usepackage{booktabs,amsmath,algorithm,algorithmic,multirow,graphicx,epstopdf} 
\newcommand{\tabincell}[2]{\begin{tabular}{@{}#1@{}}#2\end{tabular}}

\newif\ifdraft
\drafttrue 

\definecolor{orange}{rgb}{1,0.5,0}
\definecolor{violet}{RGB}{70,0,170}
\definecolor{magenta}{RGB}{170,0,170}
\definecolor{dgreen}{RGB}{0,150,0}

\usepackage[normalem]{ulem}

\ifdraft
 \newcommand{\PF}[1]{{\color{red}{\bf PF: #1}}}
 
 \newcommand{\MQ}[1]{{\color{blue}{\bf MQ: #1}}}
 
 \newcommand{\MS}[1]{{\color{dgreen}{\bf MS: #1}}}
 
 \newcommand{\WL}[1]{{\color{orange}{\bf WL: #1}}}

\else
 \newcommand{\PF}[1]{}
 
 \newcommand{\MQ}[1]{}
 
 \newcommand{\MS}[1]{}
 
  \newcommand{\WL}[1]{}
 
\fi

\newcommand{\parag}[1]{\vspace{0.5mm}\noindent {\bf #1}}

\newcommand{\mL}{\mathcal{L}}

\newcommand{\cD}{\mathcal{D}}

\newcommand{\x}{\mathbf{x}}
\newcommand{\y}{\mathbf{y}}
\newcommand{\bv}{\mathbf{v}}

\newcommand{\oursC}[0]{{\bf TASK-CYCLE}}
\newcommand{\oursCA}[0]{{\bf TASK-CYCLE-CYCADA}}

\newcommand{\oursT}[0]{{\bf TASK-CYCLE-TEMPO(L+S)}}
\newcommand{\oursTLSTM}[0]{{\bf TASK-CYCLE-TEMPO(L+S)-LSTM}}
\newcommand{\oursTLRender}[0]{{\bf TASK-CYCLE-TEMPO(L+S)-Render}}
\newcommand{\oursLT}[0]{{\bf TASK-CYCLE-TEMPO(L)}}
\newcommand{\oursST}[0]{{\bf TASK-CYCLE-TEMPO(S)}}

\usepackage{multirow}

\UseRawInputEncoding

\begin{document}
	
	\title{Unsupervised Domain Adaptation with Temporal-Consistent Self-Training for 3D Hand-Object Joint Reconstruction}

	\author{Mengshi~Qi,~\IEEEmembership{Member,~IEEE,}
		Edoardo~Remelli, 
		Mathieu~Salzmann,         
		and~Pascal~Fua,~\IEEEmembership{Fellow,~IEEE}
		\thanks{M. Qi, E. Remelli, M. Salzmann, and P. Fua are with Computer Vision Laboratory, \'{E}cole polytechnique f\'{e}d\'{e}rale de Lausanne, Lausanne CH-1015, Switzerland. (Corresponding author: Mengshi Qi, E-mail: \{firstname.lastname\}@epfl.ch). \protect\\
		}}
				
		\markboth{IEEE TRANSACTIONS ON IMAGE PROCESSING}%
		{Shell \MakeLowercase{\textit{et al.}}: Bare Demo of IEEEtran.cls for IEEE Journals}
			
			\maketitle
			

\begin{abstract}

Deep learning-solutions for hand-object 3D pose and shape estimation are now very effective when an annotated dataset is available to train them to handle the scenarios and lighting conditions they will encounter at test time. Unfortunately, this is not always the case, and one often has to resort to training them on synthetic data, which does not guarantee that they will work well in real situations. In this paper, we introduce an effective approach to addressing this challenge by exploiting 3D geometric constraints within a cycle generative adversarial network (CycleGAN) to perform domain adaptation. Furthermore, in contrast to most existing works, which fail to leverage the rich temporal information available in unlabeled real videos as a source of supervision, we propose to enforce short- and long-term temporal consistency to fine-tune the domain-adapted model in a self-supervised fashion. 

We will demonstrate that our approach outperforms state-of-the-art 3D hand-object joint reconstruction methods on three widely-used benchmarks and will make our code publicly available.

\end{abstract}


\section{Introduction}
\label{sec:introduction}

\begin{figure}[t]
	\centering
	\includegraphics[width=0.35\textwidth]{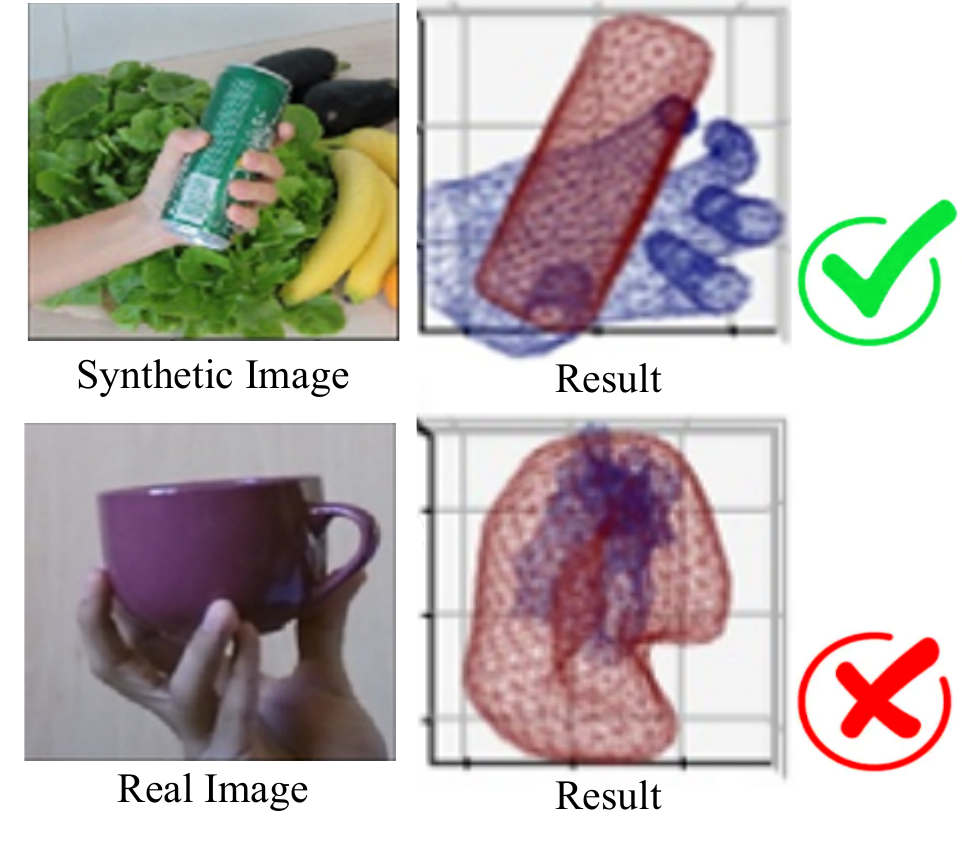}
	\vspace{-2mm}
	\caption{\small {\bf From synthetic to real data.} {\bf Top.} It is common practice to train networks on synthetic data, in this case to estimate hand-object pose and shape. {\bf Bottom.} Even though the network of~\cite{Hasson19} performs well on the synthetic data, domain shift can cause it to fail on real data.}
	\label{fig:hand_issue}
	\vspace{-2mm}
\end{figure}

 \IEEEPARstart{H}{and}-object 3D joint reconstruction is one of many computer vision problems to which convolutional neural networks have brought increasingly effective solutions~\cite{Iqbal18, Mueller18, Simon17, Zimmermann17,Spurr18}, including algorithms that can model both the hand and the object it is grasping~\cite{Hasson19,Hasson20}. However, these methods remain difficult to deploy in practice due to the lack of  large annotated datasets with ground-truth 3D hand-object pose and shape that cover a wide enough range of scenarios and lighting conditions. A common solution is therefore to train on synthetic data. Unfortunately, as shown in Fig.~\ref{fig:hand_issue}, a network trained in this fashion can easily fail when dealing with real-world images whose statistics are different from the synthetic ones. 
 

This is known as the {\it domain shift} problem and the whole field of Domain Adaption is devoted to mitigating it, ideally without requiring any additional annotations in the target domain, which is referred to as Unsupervised Domain Adaptation~(UDA)~\cite{Ganin15,Csurka17a}. In its usual formulation~\cite{Ganin15,Long16,Saito18a,Tzeng14,Long15a,Long17b,Zellinger17,Koniusz17b,Rozantsev19,Bermudez20}, one starts with a large number of annotated {\it source images} and another set of {\it target images} whose statistics are those of the images we intend to use in practice but for which we have no annotations. In our case, the source images are the synthetic ones, and the target ones are real images acquired in specific situations in which we want our software to operate. Furthermore, the complex background, various occlusion, and rapidly changing illumination in realistic environments always make it arduous to generalize the parameters of model to the different domain data. Therefore, it is necessary to make use of unlabeled target domain data to help fine-tune the model.
 
\begin{figure}[t]
	\centering
	\includegraphics[width=0.45\textwidth]{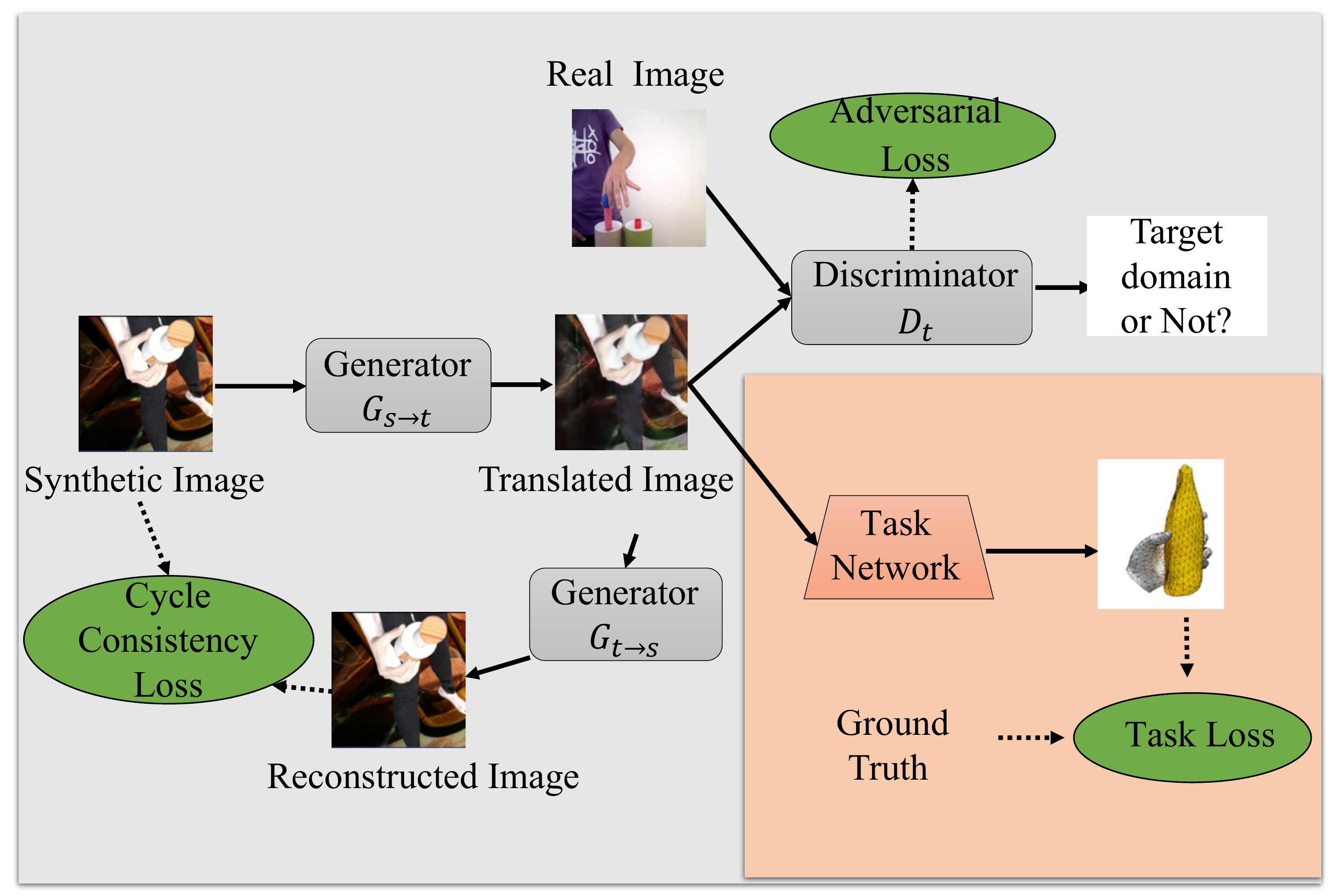}
	\vspace{-2mm}
	\caption{\small {\bf Approach.} In this work, we take synthetic images to be the source domain and real one to be the target one. In our approach, we use the CycleGAN component $G_{s\rightarrow t}$ to translate source domain images into ones that cannot be discriminated from target domain images and we use the translated images to retrain the task network.}
\label{fig:approach}
\vspace{-1mm}
\end{figure}

A promising direction to address UDA is to use a Cycle-consistent Generative Adversarial Network~(CycleGAN)~\cite{Zhu17a} to turn the annotated source domain images into images that are statistically indistinguishable from the target ones and can be used to train the network. This was demonstrated for semantic segmentation~\cite{Hoffman18}. In this work, we extend this approach for 3D hand-object reconstruction as depicted by Fig.~\ref{fig:approach}. At training time, we use a CycleGAN to translate labeled source images into target-style images. We minimize an adversarial loss to ensure the statistical similarity of the translated images and the real target images while jointly training a 3D hand-object reconstruction network to predict the correct 3D poses and shapes. We further increase performance by introducing unlabeled target video data to add an element of self-supervision by enforcing short- and long-term consistency between the predictions.

We will demonstrate that our approach outperforms state-of-the-art 3D joint hand-object reconstruction methods on three widely-used benchmarks and that our approach to UDA for hand-object reconstruction is more effective than other standard ones.

Our main contributions are summarized as follows:
\begin{itemize}
	\item We propose a 3D geometric constraints based CycleGAN model to bridge the domain gap between synthetic and realistic image domain. 
	\item We design a short-term and a long-term temporal consistency loss for self-supervised fine-tuning to make full use of unlabeled video-based real data.
	\item Extensive experiments on three public benchmarks demonstrate that our proposed approach outperforms the state-of-the-art UDA methods and baselines.

\end{itemize}


In the remainder of the paper, we first review related works on unsupervised domain adaptation and 3D hand-object reconstruction. We then present the version of our method that relies only the CycleGAN followed by the more sophisticated version that also uses the unannotated data for self-supervision. Finally, we specialize it for 3D hand-object reconstruction and present our results.

\section{Related Work}
\label{sec:rel}

In this section we briefly review existing deep learning-based approaches to jointly modeling hands and the objects they can hold in 3D. We then turn to existing domain adaptation techniques and unsupervised approaches that could be used to train networks on synthetic data, refine them using unannotated real-images, and finally infer reliable poses from real images.

\parag{Hand and Object Joint 3D Reconstruction.}
The last ten years have seen an explosion in the number of papers that use deep networks and conventional methods to model the interaction between hands and objects, which have been applied in scene understanding~\cite{qi2019attentive,qi2019ke,qi2020stc} and video analysis~\cite{qi2018stagnet,qi2020few,qi2020imitative}. Many are devoted to reconstruction from RGB-D or multi-view RGB~\cite{Ballan12,Oikonomidis11,Wan19c,Wan18a,Wan20a} input. They can be roughly classified into those that use generative methods~\cite{Hamer10,Hamer09,Oikonomidis12,Pham17,Sridhar16,Tsoli18,Tzionas15,Wan18a,Wan19c,Wan20a} and those that rely on discriminative ones~\cite{Rogez15a,Romero10,Rogez15b,Tzionas16}.  

There has been somewhat less interest for deep learning-based hand/object reconstruction from single RGB images. Unified frameworks for 3D hand-object poses estimation, interaction activity recognition, and synthetic datasets related to these tasks were introduced in~\cite{Tekin19,Hasson19,Hasson20,Armagan20a,Brahmbhatt20a,Hampali20,Doosti20a,Mueller18}. The resulting methods perform well on the kind of synthetic data for which they have been trained but not so well on real-images, as shown in the example of Fig.~\ref{fig:hand_issue} because of the domain shift. Our work differs from an approach such as that of~\cite{Mueller18} in that we directly use hand-object task loss to supervise the CycleGAN and train it jointly with the hand-modeling network.

\parag{Domain Adaptation.}
In recent years, the dominant approach to tackling the domain shift issue has been to learn a domain-invariant representation~\cite{Csurka17a}.  One way is to map source data and target data into a latent space by minimizing the Maximum Mean Discrepancy~\cite{Gretton07} or matching second and higher order statistics~\cite{Koniusz17a,Sun17b,Sun16c} between the source and target domains~\cite{Long16,Saito18b,Tzeng14,Long15a,Long17b,Zellinger17,Koniusz17b,Yan17}. Class labels~\cite{Saito18b} and anchor points~\cite{Hausser17,Shkodrani18} can also be used. Another way consists of leveraging adversarial training to learn source and target features that are indistinguishable~\cite{Ganin15,Bousmalis17,Hong18a, Vu19, Li19b, Hoffman18, Tzeng15, Tzeng17, Goodfellow14,Ganin16,Hu18c,Zhang18c}. Beyond image recognition, this has been applied to semantic segmentation~\cite{Chen18d,Hoffman18,Hong18a} and active learning~\cite{Su19}, and has been the focus of our own recent work, in which the source and target data pass through networks with the same architecture but different weights~\cite{Rozantsev19,Bermudez20}. 

An interesting alternative to these two dominant approaches consists of translating the images from one domain to  the other using a CycleGAN~\cite{Zhu17a}, thus making it possible to adapt the images directly at the pixel level. To this end, Cycada~\cite{Hoffman18} uses a CycleGAN to translate source images into  target-like images that can be used along with the annotations to train a network that performs the desired task to operate in the target domain. Specifically, the CycleGAN and the task network can be trained jointly, which has inspired our own approach. Here, however, we translate this idea to the geometric task of reconstructing 3D hand and object from a single image. 

\parag{Self-Supervised Learning.}
Another approach to reducing the required amount of annotated data is to rely on self- or semi-supervised learning~\cite{Zhang2016d,Larsson16,Vondrick18,Noroozi16,Doersch15,Gidaris18}. To provide a useful self-supervisory signal, current methods rely on image color~\cite{Zhang2016d,Larsson16,Vondrick18}, relative position of image patches~\cite{Noroozi16,Doersch15}, random image rotations~\cite{Gidaris18,Zhai19}, missing part of image~\cite{Pathak16}, multi-view consistency and depth information~\cite{Godard17}, depth hints from left and right cameras~\cite{Sterzentsenko19,Watson19}, temporal consistency~\cite{Lee17b,Misra16b,Vondrick18,Zhou17c}, multi-domain invariant representation~\cite{Feng19}, and image clustering~\cite{Larsson19}. In this work, we show that temporal consistency effectively complements domain adaptation and makes it possible to leverage unannotated video sequences for hand and object 3D reconstruction.

\section{Approach}
\label{sec:app}
						
In this work, we propose a new UDA framework that exploits a generative adversarial network (CycleGAN) to translate source synthetic images into realistic target-like images for training purposes. We use the translated source images along with the corresponding annotations to train the target network to perform the task of interest in the target domain. Hence, at inference time, we do not need the CycleGAN anymore and the network can operate on unmodified target images. The key to making this scheme work properly is to impose a task loss, \ie, 3D geometric preservation constraint while training the CycleGAN model so that the translated images retain enough key information for the target network to work well when trained using them. Furthermore, we propose to self-supervised fine-tune the task network by adding a ConvLSTM~\cite{Shi15} layer using unlabeled target video data by enforcing long-term and short-term temporal consistency of its predictions.

In this section, we first formalize our UDA problem. We then describe our proposed CycleGAN-based model, and our self-supervised training strategy with the temporal consistency losses.

\subsection{Formalization} 
\label{sec:formal}

We use the encoder-decoder architecture of~\cite{Hasson19}, which we will refer to as the task network $F$. It takes images as input and returns 3D models for the hand and the object it interacts with. These models include coordinate vectors for vertices of the surface meshes representing the hand and the object surface along with hand pose and shape parameters. We will refer to them collectively as {\it hand-object vectors}, which will be described in more detail in Section~\ref{sec:implementation}.


Let us consider a source domain  $\mathcal{D}_s=\{\x_s,\y_s\}^{N_s}_{i=1}$ with $N_s$ images and corresponding ground-truth hand-object vectors, and a target domain $\mathcal{D}_t=\{\x_t\}^{N_t}_{i=1}$ in which we only have $N_t$  images. We can use $\mathcal{D}_s$ to train a source network $F_s$ and our goal is to learn new network weights for the target network $F_t$ using only $\mathcal{D}_s$ and $\mathcal{D}_t$ so that the network operates effectively in the target domain.

To perform domain adaptation, we introduce a CycleGAN that comprises two generator networks $G_{s \rightarrow t}$ and $G_{t \rightarrow s}$ that translate source domain images into target domain ones and target domain images into source ones, respectively, along with a discriminator network $D_t$ that attempts to discriminate translated source images from true target images. 
Fig.~\ref{fig:approach} depicts the role of these networks. 


\subsection{Joint Training the CycleGAN and Task Networks}
\label{sec:joint}


We iteratively train the generator network $G_{s \rightarrow t}$ and task network $F_t$ so as to maximize the performance of the latter on target images. Let $F_t^{(0)} = F_s$ be the task network pre-trained on the source data and let $F_t^{(k-1)}$ be the task network after iteration $k-1$. At iteration $k$, we feed a source image $\x_s$ to the generator network $G_{s \rightarrow t}$, which outputs a translated source image $G_{s \rightarrow t}(\x_s)$. In turn, it serves as input to the task network $F_t^{(k-1)}$ to add a task-related constraint during training CycleGAN. To update the weights of  $F_t$, $D_t$, and $G_{s\rightarrow t}(\x_s)$, we minimize a loss that comprises the three loss functions depicted by green ellipses in Fig.~\ref{fig:approach}. They are defined as follows. The first two are standard while the third one is specific to our approach.

\parag{Adversarial Loss:  $\mathcal{L}_{\rm{adv}}$}. It favors translated source images that are statistically indistinguishable from target ones. We take it to be
\begin{small}
 \begin{equation}
 \label{eq:cycle1}
 \mathcal{L}_{\rm{adv}}=\mathbb{E}_{{\x_t \in \cD_t}}[\log D_t(\x_t)]+\mathbb{E}_{{\x_s \in \cD_s}}[\log(1\hspace{-1mm}-\hspace{-1mm}D_t(G_{s\rightarrow t}(\x_s))] \; .
 \end{equation}
 \end{small}

\parag{Cycle-consistency Loss: $\mathcal{L}_{\rm{cyc}}$.} It ensures that the translated images can be translated back into the original ones. We write it as
 \begin{small}
 \begin{equation}
 \label{eq:cycle2}
 \mathcal{L}_{\rm{cyc}}=\mathbb{E}_{{\x_s \in \cD_s}}[\|G_{t\rightarrow s}(G_{s\rightarrow t}(\x_s))-\x_s\|_1] \; .
 \end{equation}
 \end{small}
 
 \parag{Task Loss: $\mathcal{L}_{\rm{task}}$.} It is the usual supervised loss used to train the task network given samples and corresponding labels. It is application specific and we describe it Section~\ref{sec:implementation}.

\begin{figure}[t]
	\centering
	\includegraphics[width=0.45\textwidth]{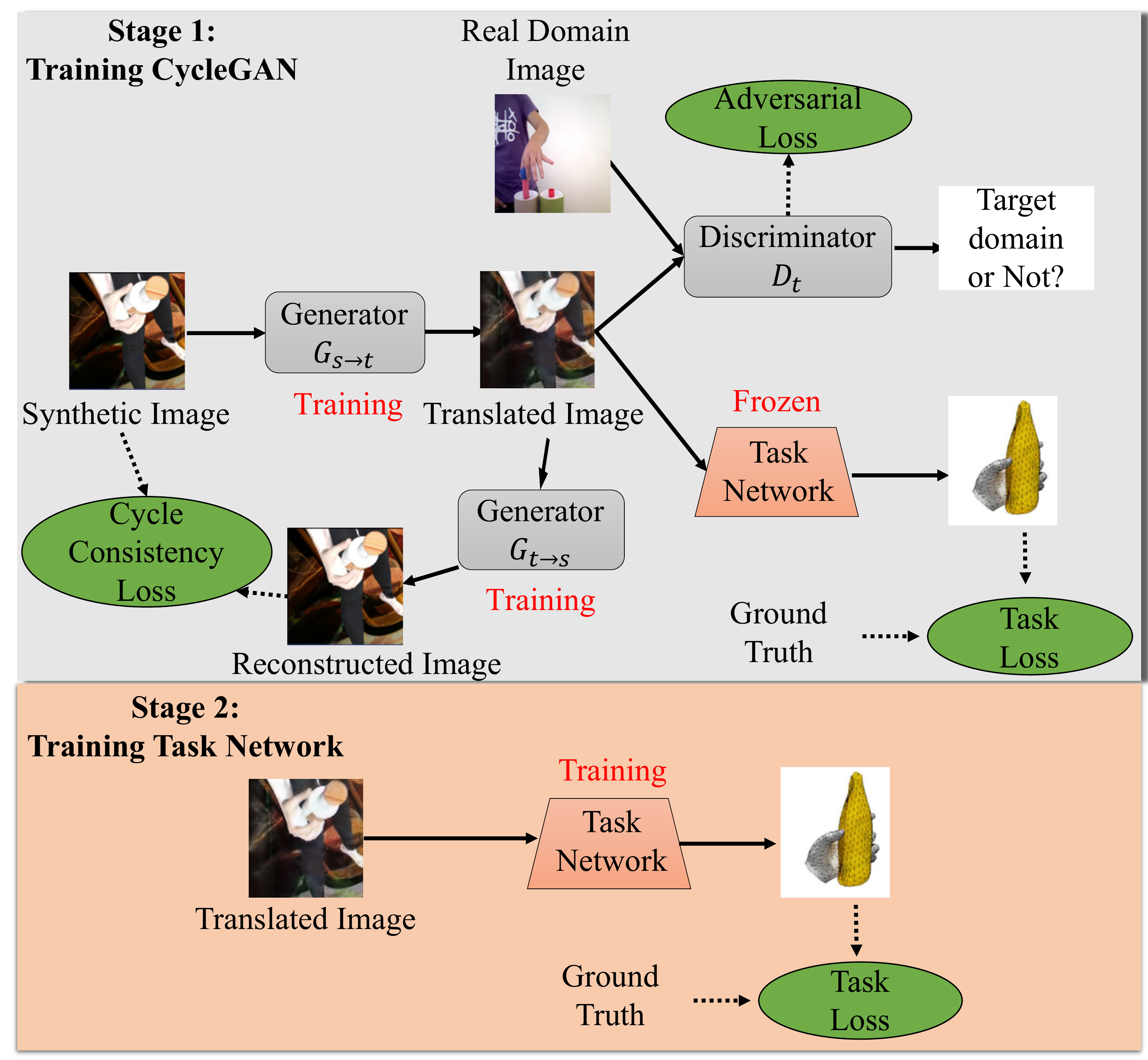}
	\vspace{-1mm}
	\caption{\small {\bf Two-Stage Training.} We alternatively train the generators with the task network frozen and then the task network with the generators frozen.}
\label{fig:training}
\vspace{-3mm}
\end{figure}

\parag{Optimization Strategy:}  We start with task network $F_t^{(0)}$ and generator networks  $G_{s \rightarrow t}$ and $G_{t \rightarrow s}$ trained in the usual manner. In theory, we could then simultaneously train them further by minimizing the joint objective function 
\begin{small}
\begin{equation}
\mL_{\rm{UDA}}=\mathcal{L}_{\rm{adv}} +\lambda_{\rm{cyc}}\mathcal{L}_{\rm{cyc}}+\lambda_{\rm{task}}\mathcal{L}_{\rm{task}} \; 
\label{eq:cycle4}
\end{equation}
\end{small}
with respect to their weights, where $\lambda_{cyc}$ and $\lambda_{\rm{task}}$ are scalar coefficients. In practice, however, the full architecture depicted by Fig.~\ref{fig:approach} is too heavy for this. Instead, we alternatively optimize the weights of $F^{(k-1)}$ into those of  $F^{(k)}$ with those of the generators fixed and then the weights of the generators with those of the task network frozen. Fig.~\ref{fig:training} depicts this process. We have found empirically that the process converges better when we give more weight to the task loss than the cycle loss. We therefore set  $\lambda_{cyc}=0.1$ and $\lambda_{\rm{task}}=1$. 

\subsection{Implementation}
\label{sec:implementation}


\begin{figure}[t]
	\centering
	\begin{tabular}{cc}
	\includegraphics[width=0.45\textwidth]{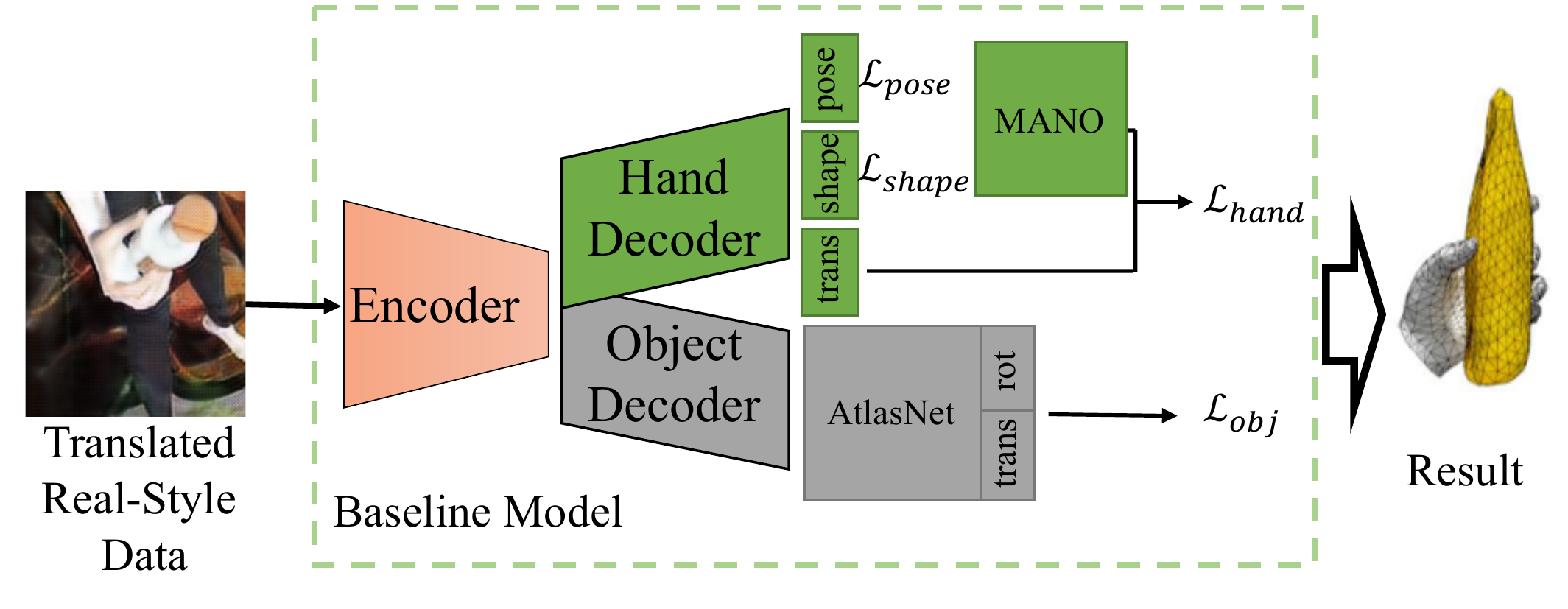}
	\end{tabular}
	\vspace{-1mm}
	\caption{\small {\bf Task network}. We use the encoder-decoder architecture of~\cite{Hasson19}. It takes as input an image and returns 3D mesh models for both the hand and the object.}
	\label{fig:task_nets}
    \vspace{-1mm}
\end{figure}

We take the task network to be the encoder-decoder architecture of~\cite{Hasson19} depicted by Fig.~\ref{fig:task_nets} and use its publicly available implementation.  A ResNet-18~\cite{He16a} pre-trained on ImageNet~\cite{Deng09} serves as the image encoder. The {\it hand decoder} predicts the shape and pose parameters,  $\theta_{H}$ and $\beta_{H}$, of MANO~\cite{Romero17}, which are used to compute the vertex coordinates $\bv_{\rm hand}$ of the mesh describing the hand. The {\it object decoder} outputs 3D pose parameters that are combined with the output of an AtlasNet~\cite{Groueix18a} network to yield the vertex coordinates of a 3D mesh describing the object  $\bv_{\rm obj}$.  Principal Component Analysis~(PCA) is used to keep the dimensions of $\theta_{H}$ to 15 and that of $\beta_{H}$ to 10. Hence, the task loss function  $\mathcal{L}_{\rm{task}}$ introduced in Section~\ref{sec:joint} can be written as
\begin{small}
\begin{align}
\mathcal{L}_{\rm{task}}&=\mathcal{L}_{\rm obj}+\lambda_{h}\mathcal{L}_{\rm hand}+\lambda_{s}\mathcal{L}_{\rm shape}+\lambda_{\rm p}\mathcal{L}_{\rm pose}+\lambda_{\rm c}\mathcal{L}_{cont} \; , \nonumber \\
\mathcal{L}_{\rm obj}&=\sum\|\hat{\bv}_{\rm obj}-\bv_{\rm obj}^{gt}\|^2_2 \; , \nonumber \\
\mathcal{L}_{\rm hand}&=\sum\|\hat{\bv}_{\rm hand}-\bv_{\rm hand}^{gt}\|^2_2 \; , 	\label{eq:taskLoss} \\
\mathcal{L}_{\rm shape}&=\|\beta_H\|^2_2 \; , \nonumber\\
\mathcal{L}_{\rm pose}&=\|\theta_H\|^2_2 \; , \nonumber\\
\mathcal{L}_{\rm cont}&=\mathcal{L}_{\rm repulsion}+\mathcal{L}_{\rm attraction} \; , \nonumber
\end{align}
\end{small}
where $\bv_{\rm hand}^{gt}$, $\bv_{\rm obj}^{gt}$ and $\hat{\bv}_{\rm hand}^{gt}$, $\hat{\bv}_{\rm obj}^{gt}$ are ground-truth and estimated values, respectively. $\mathcal{L}_{\rm repulsion}$ and $\mathcal{L}_{\rm attraction}$ are additional loss terms designed to guarantee that the hand does not penetrate into the object but actually touches it when appropriate. For a more complete description we refer the interested reader to the original publication~\cite{Hasson19}. The model is implemented within the PyTorch framework and all experiments were run on one NVIDIA Tesla V100 GPU. We use the Adam optimizer~\cite{Kingma15} to train our model for 200 epochs with an initial learning rate of $5\times10^{-5}$ and a batch size of 16. 


\subsection{Adding Temporal Consistency}
\label{sec:tempo}

The method of Section~\ref{sec:joint} treats all unannotated images as individual images as opposed to elements of a video sequence in which motions are continuous. This fails to exploit the fact that, even though the 3D poses in consecutive images are initially unknown, the network predictions should be consistent across consecutive images and without brutal jumps. In this section, we use this fact to define a new loss term that enforces temporal consistency both over the short and the long term. 

Formally, let $\cD_t=\{\x^T\}_{T=1}^{N_t}$ be a sequence of $N_t$ consecutive unlabeled target domain images from a realistic video. For each one, the task network $F_t$ outputs a description of the hand and object at time $T$ in the form of  an output vector $\bv^T$ that encodes the vertex coordinates for the hand and object meshes, along with hand pose and shape vectors $\theta_{H}^T$ and $\beta_{H}^T$, as described in Section~\ref{sec:formal}.



\begin{figure}[t]
	\centering
	\includegraphics[width=0.45\textwidth]{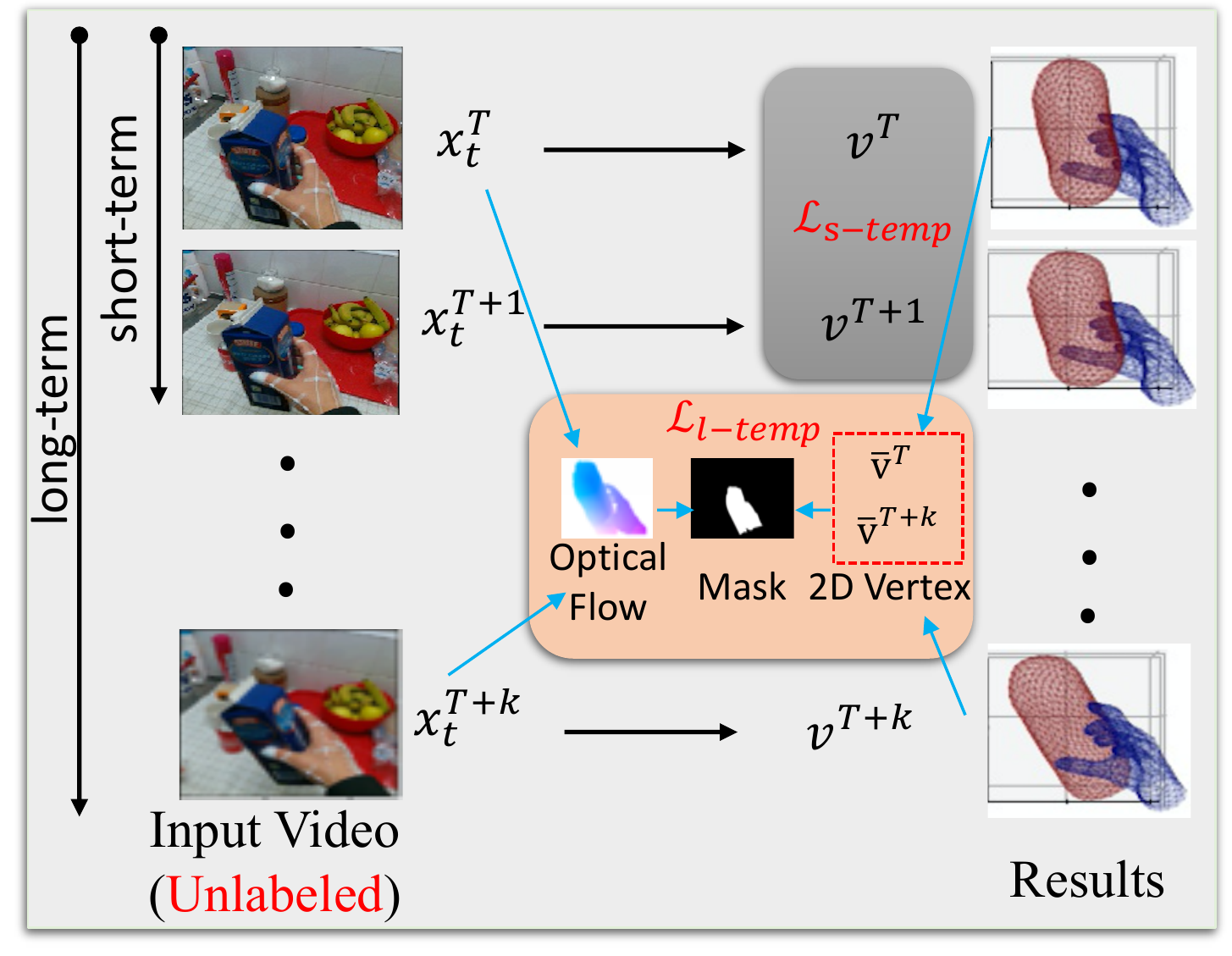}
	\vspace{-1mm}
	\caption{\small {\bf Temporal consistency loss.} Given an unannotated video sequence, we compute a smoothness-based short-term consistency loss in neighboring frames $x_t^T$ and $x_t^{T+1}$ and a long-term one between frames  $x_t^T$ and $x_t^{T+k}$ with $k>1$. The latter depends on the consistency between the predicted projected displacement between mesh vertices and the optical flow between the two images.}
	\label{fig:temproal_cons}
	\vspace{-1mm}
\end{figure}

\parag{Short-Term Temporal Consistency.} 
We require the predictions at consecutive time instants to be similar by introducing the short-term loss
\begin{small}
	\begin{equation}
	\label{eq:short}
	\begin{split}
	\mathcal{L}_{\rm{s-temp}}&= \sum_T(\mathcal{L}_{\rm{mano}}+\lambda_{s-temp}\|\bv^{T+1}-\bv^T\|^2_2); \\
	\mathcal{L}_{\rm{mano}}&=\|\beta^{T+1}_H-\beta^T_H\|^2_2+\lambda_{s-temp}\|\theta^{T+1}_H-\theta^T_H\|^2_2   \; ,
	\end{split}
	\end{equation}
\end{small}
where the summation occurs over a video sequence. As the hand shape parameters remain constant over such a sequence whereas the pose parameter change, we impose a stronger constraint on the former and set $\lambda_{s-temp}=0.01$. Note that we penalize the change in vertex coordinates before applying the roto-translation that maps object-centered vertex coordinates into camera-centered ones with AtlasNet. 

\parag{Long-Term Temporal Consistency.}
Because the images at time $T$ and $T+1$ are different but still similar, $\mathcal{L}_{s-temp}$ provides a supervisory signal but one that is not as informative than the one that could be obtained from images that are further in time. To this end and as in~\cite{Hasson20}, we use an independent estimate~\cite{Hu16} of the optical flow $\mathcal{O}_{T}^{T+k}$ between images acquired as time $T$ and $T+k$ to enforce temporal consistency on the vertices  predicted at time $T$ and $T+k$, \ie, $\bv^{T}$ and $\bv^{T+k}$, respectively. Then the displacement between the 2D projections of these vertices, $\bar{\bv}^{T}$ and $\bar{\bv}^{T+k}$, should be consistent with the corresponding optical flow values. We therefore take the long-term temporal consistency loss to be
 \begin{small}
 	\begin{align}
	\mathcal{L}_{\rm{l-temp}}         &= \sum_n \mathcal{L}_{l-temp}^{T,nk}  \label{eq:long_temp1} \\
 	\mathcal{L}_{\rm{l-temp}}^{T,nk} &= \frac{1}{N_v}\sum_{i=1}^{N_v} M^T(\bar{\bv}^{T}_i) \|(\mathcal{O}_T^{T+k}(\bar{\bv}^{T}_i) -(\bar{\bv}^{T+k}_i-\bar{\bv}^{T}_i))\|_2^2 \; ,	\nonumber
 	\end{align}
 \end{small}
where $N_v$ is the total number of vertices and $M^T$ is a binary visibility mask at time $T$. It is computed by performing an optical flow forward-backward consistency check as in~\cite{Hasson20,Wang18f}. In practice, we take $T=1$ and $k=5$.

\parag{Optimization Strategy:} 
We first train the task network $F_t$ on individual images as described in Section~\ref{sec:joint}. We then fine-tune the network in a self-supervised fashion by minimizing 
\begin{small}
	\begin{equation}
	\mathcal{L}_{\rm{self}}=\mathcal{L}_{\rm{s-temp}}+\lambda_{l}\mathcal{L}_{\rm{l-temp}},
	\label{eq:fullLoss}
	\end{equation}
\end{small}
\noindent where $\lambda_{l}=0.1$. During this minimization, the generator networks are frozen and only the task network weights are refined. In fact, we have found it advantageous to also freeze the first layers of the task network and to only refine the final layer, that is, a ConvLSTM~\cite{Shi15} layer with the hidden size of 512 added between the encoder and the last fully connected layer. \

\section{Experiments}
\label{sec:exp}

In this section, we first introduce the baselines against which we compare several variants of our approach. We then present the experimental results and analysis in terms of 3D hand-object joint reconstruction. 

\subsection{Variants and Baselines}
\label{sec:baseline}

\begin{figure}[t]
	\centering
	\includegraphics[width=0.95\columnwidth]{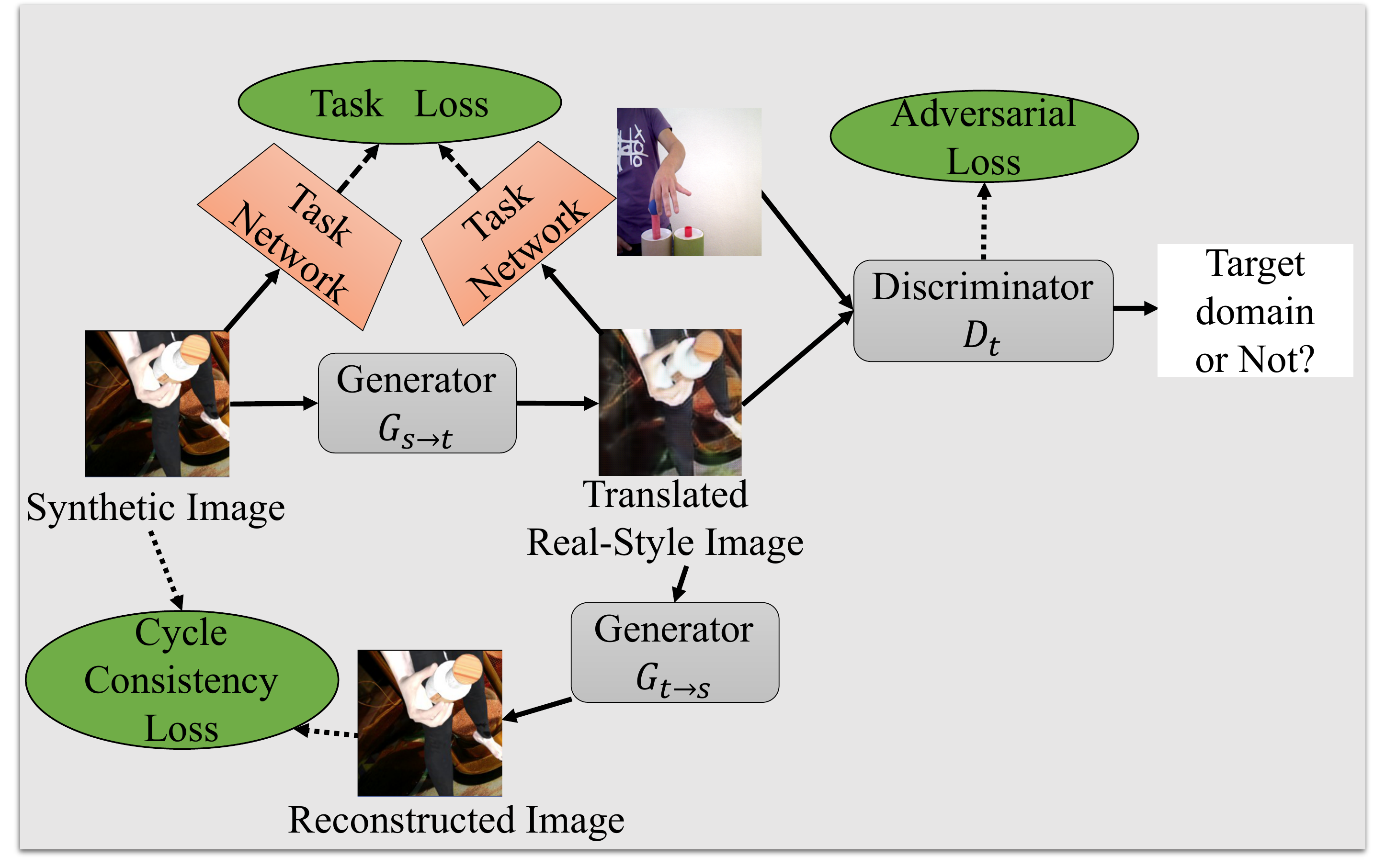}
	\vspace{-1mm}
	\caption{\small {\bf Cycada like variant of our approach.} The task loss is now used to compare the output of the network on the input synthetic data and the translated images.}
	\label{fig:cycada}
	\vspace{-1mm}
\end{figure}

To compare against other approaches to UDA, we use the following baselines:
\begin{itemize}

   \item {\bf No DA}:  Running the task network~\cite{Hasson19}  trained on synthetic data and test on real data without any domain adaptation. 
   \item {\bf UDA w/ CycleGAN~\cite{Zhu17a}}: Using a standard CycleGAN to translate synthetic images and use the translated images to train the task network.
   \item {\bf UDA w/ FDA~\cite{Yang20c}}:    Using a simple Fourier transform to translate the image.
   \item {\bf UDA w/ ADAA~\cite{Tzeng17}}: Performing adversarial discriminative domain adaptation on the encoder of the task network to learn domain-invariant representations. 
   \item {\bf UDA w/ DANN~\cite{Ganin15}}: Adding the gradient reversal layers to the task network to learn domain-invariant representations.
\end{itemize}
Moreover, we compare them to the following variants of our approach:
\begin{itemize}
	\item {\oursC}:              Iterative training of the generators and task network as described in Section~\ref{sec:joint}.
	\item {\oursT}:              Adding long-term and short-term temporal consistency as described in Section~\ref{sec:tempo}.
	\item {\oursST}:            Adding only  short-term temporal consistency as described in Section~\ref{sec:tempo}.
	\item {\oursLT}:             Adding only  long-term temporal consistency as described in Section~\ref{sec:tempo}.
	\item {\oursTLSTM}:     Replacing the ConvLSTM layer in the task network by an LSTM layer during self-supervised fine-tuning.	
	\item {\oursTLRender}: Computing the long-term consistency using an approach similar to~\cite{Hasson20}, that is, first using Neural Render~\cite{Kato18} to render the hand and object meshes and then estimating the consistency of the rendered images and the optical flow. 
	\item {\oursCA}:           Instead of using the task loss to compare the output of the network to the ground truth, using it to compare to the images {\it before} translation, as in~\cite{Hoffman18} and as shown in Fig.~\ref{fig:cycada}.
\end{itemize}


\subsection{Datasets}

We evaluate our framework on one large-scale synthetic image dataset~\emph{ObMan}~\cite{Hasson19}, and two widely-used real image datasets---\emph{First-person hand benchmark}~\cite{Garcia18} and~\emph{Hands in action dataset}~\cite{Tzionas16}---that we describe in more details below.
			
\noindent{\bf ObMan Dataset~\cite{Hasson19}:}~ObMan is the largest fully labeled synthetic image dataset describing hands grasping objects. It contains 2.7k objects models of eight categories from the ShapeNet dataset~\cite{Chang15}~(i.e., bottles, bowls, cans, jars, knifes, cellphones, cameras and remote controls), and grasping actions generated by the GraspIt software~\cite{Miller04}. Furthermore, all hand models are fitted with MANO~\cite{Romero17} to grasp a provided object mesh, and the hands are rendered by SMPL~\cite{Romero17}. The body and object textures were taken from SURREAL~\cite{Varol17} and ShapeNet~\cite{Chang15}. Each scene was rendered using Blender~\cite{Blender17} with a random lighting and in front of background images randomly sampled from LSUN~\cite{Yu15b} and ImageNet~\cite{Deng09}. Following~\cite{Hasson19}, we select 141k frames and 6K frames as training set and test set, respectively.
			
\noindent{\bf First-person hand benchmark~(FHB)~\cite{Garcia18}:}~FHB is an RGB-D video-based dataset of first-person hand-object interactions. The data is fully annotated with hand's and object's 6D pose, 3D location and mesh model. Following the setting in~\cite{Hasson19}, to focus on hand-object interaction, we only select the frames in which the distance between the manipulating hand and the manipulated object is less than 1cm, and choose three categories of objects, i.e.,~\emph{salt}, \emph{juice carton} and \emph{liquid soap}. This corresponds to a subset of FHB with 8420 training frames and 9103 testing ones.
			
\noindent{\bf Hands in action dataset~(HIC)~\cite{Tzionas16}:}~Following the experimental setting in~\cite{Hasson19}, we use four video sequences from HIC, and employ two sequences~(251 frames) as training set and two others~(307 frames) as test. All frames feature interactions between one hand and a sphere or a cube. In our experiments, we choose the frames in which the minimal distance between the hand and the vertices of the manipulated object is below 5mm. Furthermore, the HIC dataset provides hand and object meshes, to which we fit the MANO~\cite{Romero17} model for supervision with dense 3D points. 


\begin{table*}[!t]
\caption{\small {\bf Comparative results} on FHB and HIC datasets. We provide both mean values and variances over several trials. The best results are shown in bold.}
\vspace{-1mm}		
	\centering
		\resizebox{0.995\linewidth}{!}{
		\begin{tabular}{lcccccccccc}
			\toprule
			\multirow{2}{*}{Methods}    &\multicolumn{5}{c}{FHB Dataset} &\multicolumn{5}{c}{HIC Dataset} \\
			\cmidrule(lr){2-6}\cmidrule(lr){7-11} &\tabincell{c}{Hand\\Error}&\tabincell{c}{Object\\Error}&\tabincell{c}{Maximum\\Penetration}&\tabincell{c}{Simulation\\Displacement}&\tabincell{c}{Intersection\\volume}&\tabincell{c}{Hand\\Error}&\tabincell{c}{Object\\Error}&\tabincell{c}{Maximum\\Penetration}&\tabincell{c}{Simulation\\Displacement}&\tabincell{c}{Intersection\\volume}    \\  \midrule
			No DA-only source                        &32.7 $\pm$ 0.7  &2079.2 $\pm$ 58.7  &13.7 $\pm$ 0.6   &49.2 $\pm$ 2.2    &19.7 $\pm$ 0.3  &33.1 $\pm$ 0.7   &2150.0 $\pm$ 66.5   &23.7 $\pm$ 0.9    &62.5 $\pm$ 2.0   &36.7 $\pm$ 0.3     \\
			UDA~w/~CycleGAN~\cite{Zhu17a}    &26.9 $\pm$ 0.6  &2057.3 $\pm$ 61.5  &10.9 $\pm$ 0.6  &45.1 $\pm$ 2.0    &16.1 $\pm$ 0.3   &28.0 $\pm$ 0.6  &2140.7 $\pm$ 63.5   &21.5 $\pm$ 0.8    &59.1 $\pm$ 1.6   &34.9 $\pm$ 0.3     \\ 
			UDA~w/~FDA~\cite{Yang20c}           &27.2 $\pm$ 0.7  &2062.3 $\pm$ 62.2  &11.7 $\pm$ 0.7   &46.6 $\pm$ 1.7  &16.7 $\pm$ 0.3  &28.9 $\pm$ 0.7  &2147.5 $\pm$ 65.0  &22.9 $\pm$ 0.6   &61.2 $\pm$ 1.9  &35.7 $\pm$ 0.5    \\ 
			UDA~w/~ADDA~\cite{Tzeng17}        &28.1 $\pm$ 0.6  &2064.5 $\pm$ 57.5  &11.9 $\pm$ 0.6  &47.5 $\pm$ 1.8  &17.3 $\pm$ 0.2  &29.8 $\pm$ 1.1  &2149.3 $\pm$ 61.2  &23.5 $\pm$ 0.7  &62.2 $\pm$ 2.1  &36.3 $\pm$ 0.4     \\ 
			UDA~w/~DANN~\cite{Ganin15}         &27.8 $\pm$ 0.8  &2064.0 $\pm$ 59.0  &11.5 $\pm$ 0.5   &47.1 $\pm$ 1.9  & 16.9 $\pm$ 0.3  &29.3 $\pm$ 1.0  &2149.0 $\pm$ 62.5  &23.2 $\pm$ 0.8   &61.7 $\pm$ 2.2  &35.9 $\pm$ 0.5     \\   \midrule
			\oursC                             &25.2 $\pm$ 0.6  &2054.0 $\pm$ 58.0   &10.1 $\pm$ 0.5       &44.1 $\pm$ 1.8    &15.5 $\pm$ 0.2   &26.5 $\pm$ 0.6   &2137.3 $\pm$ 61.0   &20.1 $\pm$ 0.6    &57.7 $\pm$ 1.6   &33.7 $\pm$ 0.2      \\
			\oursT                              &{\bf 24.3 $\pm$ 0.5}  &{\bf 2052.6 $\pm$ 60.5}   &{\bf 9.5 $\pm$ 0.6}       &{\bf 43.1 $\pm$ 1.6}    &{\bf 15.0 $\pm$ 0.2}   &{\bf 25.3 $\pm$ 0.5}   &{\bf 2135.2 $\pm$ 62.2}   &{\bf 18.8 $\pm$ 0.7}    &{\bf 56.6 $\pm$ 1.7}  &{\bf 33.0 $\pm$ 0.3}      \\    \midrule
			\oursCA                           &25.5 $\pm$ 0.5  &2054.3 $\pm$ 61.9    &10.3 $\pm$ 0.6     &44.3 $\pm$ 2.1    &15.6 $\pm$ 0.3   &26.7 $\pm$ 0.6    &2137.7 $\pm$ 63.9    &20.4 $\pm$ 0.8  ß  &58.0 $\pm$ 2.0    &33.9 $\pm$ 0.4       \\ 
			\oursST                            &24.9 $\pm$ 0.2  &2053.3 $\pm$ 56.3   &9.9 $\pm$ 0.3       &43.9 $\pm$ 1.5    &15.5 $\pm$ 0.1  &26.3 $\pm$ 0.3   &2136.7 $\pm$ 60.5   &19.8 $\pm$ 0.5    &57.5 $\pm$ 1.5   &33.6 $\pm$ 0.2      \\  
			\oursLT                           &24.7 $\pm$ 0.5  &2053.0 $\pm$ 57.6   &10.0 $\pm$ 0.5       &43.7 $\pm$ 1.6   &15.3 $\pm$ 0.3   &26.1 $\pm$ 0.5   &2136.2 $\pm$ 63.2   &19.6 $\pm$ 0.6    &57.3 $\pm$ 1.6   &33.5 $\pm$ 0.3       \\  
			\oursTLSTM                   &24.5 $\pm$ 0.5  &2052.7 $\pm$ 59.0   &9.7 $\pm$ 0.5       &43.5 $\pm$ 1.6    &15.1 $\pm$ 0.2   &25.8 $\pm$ 0.6   &2135.5 $\pm$ 63.0   &19.5 $\pm$ 0.6    &57.0 $\pm$ 1.8   &33.2 $\pm$ 0.3      \\  
			\oursTLRender              &24.6 $\pm$ 0.6  &2052.5 $\pm$ 60.5  &9.7 $\pm$ 0.6        &43.4 $\pm$ 2.0    &15.2 $\pm$ 0.3  &25.7 $\pm$ 0.6   &2135.3 $\pm$ 64.7   &19.3 $\pm$ 0.7  &56.9 $\pm$ 1.8  &33.2 $\pm$ 0.5      \\  
			\bottomrule	   
	\end{tabular}}	
	\label{tab:performance_hand_1}
\end{table*}




\subsection{Metrics}

For comparison purposes, we use the same metrics as in~\cite{Tzionas16,Zimmermann17,Hasson19}:

\parag{Hand Reconstruction Error.} We compute the mean end-point error~(mm) across 21 joints as in~\cite{Zimmermann17}. 

\parag{Object Reconstruction Error.} We measure the symmetric Chamfer distance~(mm) between sampled vertices on the ground-truth mesh and points on the predicted mesh as in~\cite{Groueix18a,Hasson19}. The mean per vertex error is therefore the number we report over 642, the total number of vertices. 

\parag{Contact Quality.} We compute the~\emph{penetration depth} and~\emph{intersection volume} between reconstructed hands and objects. To this end, we use a voxel size of 0.5cm to measure the intersection volume, and the penetration depth is the maximum distance from the sampled points on the hand mesh to the surface of the object mesh. 

\parag{Simulation displacement.}  We run the simulation environment introduced in~\cite{Tzionas16} that returns a measure of the physical plausibility of our estimates. We refer the interested reader to its full description in the original paper.

\subsection{Results}

\begin{figure}[t]
	\centering
	\includegraphics[width=0.45\textwidth]{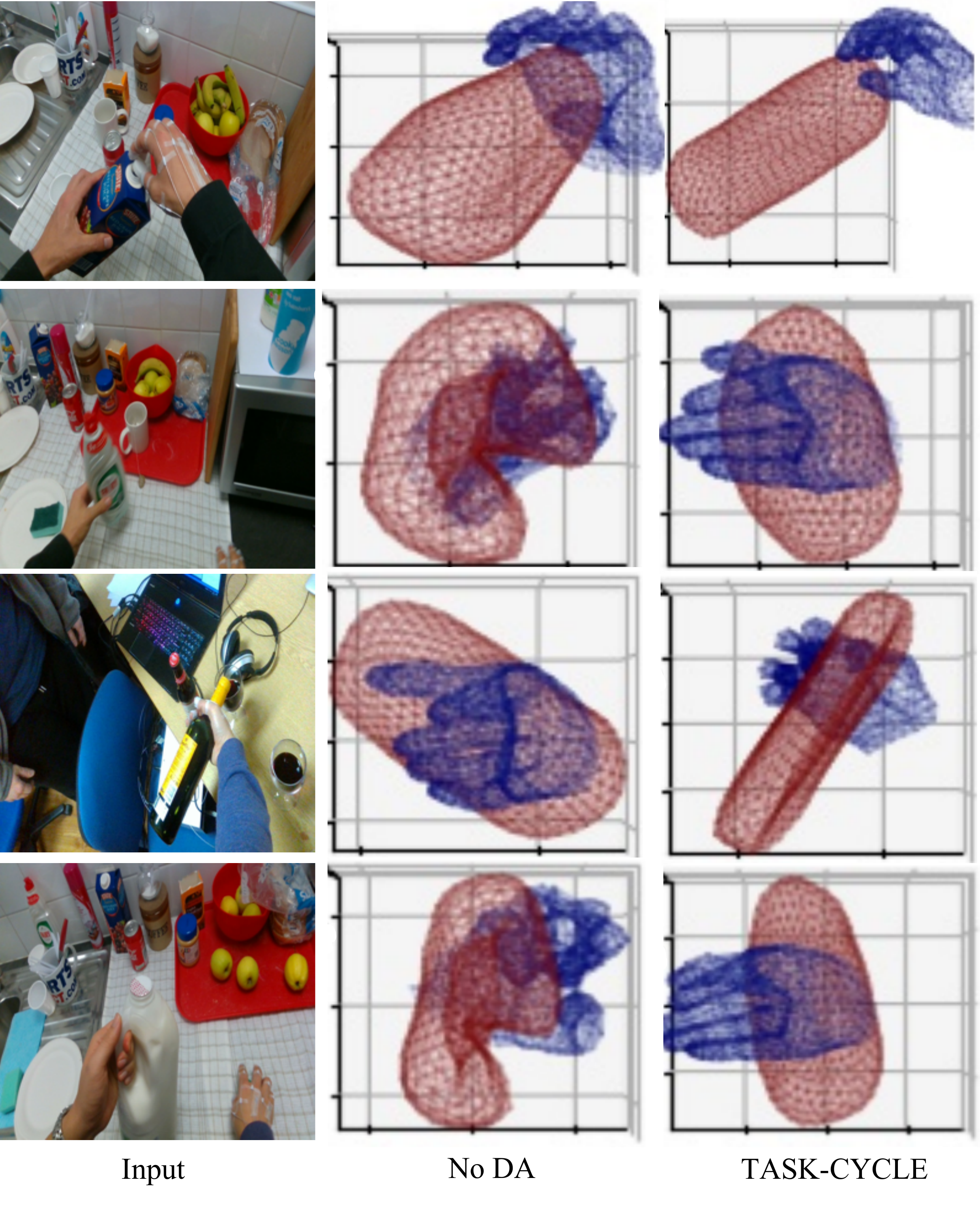}
	\vspace{-1mm}
	\caption{\small {\bf Qualitative results on single images} from the FHB dataset. The hand pose recovered by our system is correct in the first three rows but wrong in the fourth, which can be attributed to the fact that no such hand pose exists in the training set.}
	\label{fig:result_single}
	\vspace{-1mm}
\end{figure}

\begin{figure*}[t]
	\centering
	\includegraphics[width=0.85\textwidth]{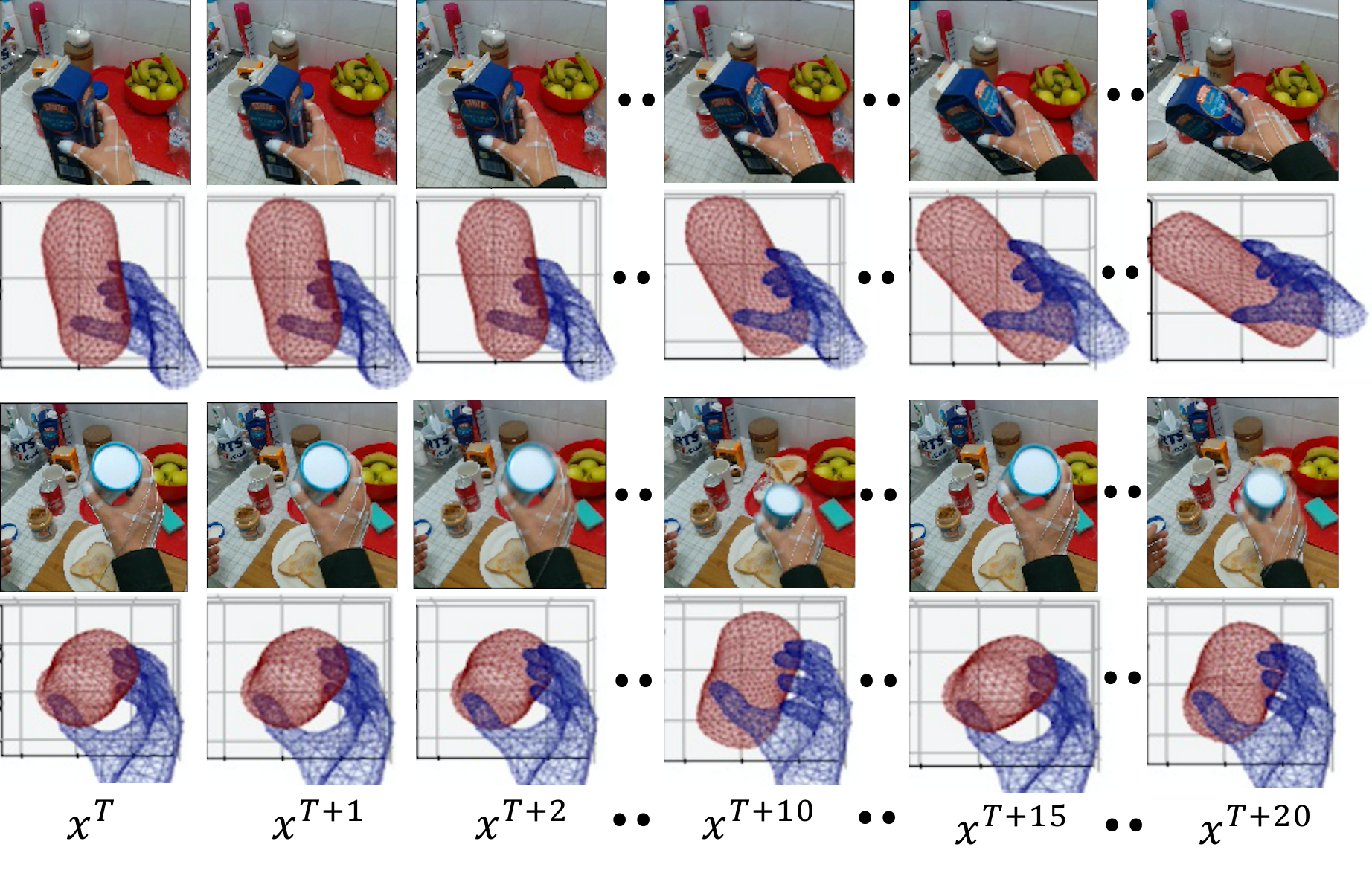}
	\vspace{-2mm}
	\caption{\small {\bf Qualitative results on two sequences} from the FHB dataset. The temporal loss was used to train the network but, at inference time, each image is processed individually.}	\label{fig:result_temporal}
\end{figure*}

\begin{figure}[t]
	\centering
	\includegraphics[width=0.45\textwidth]{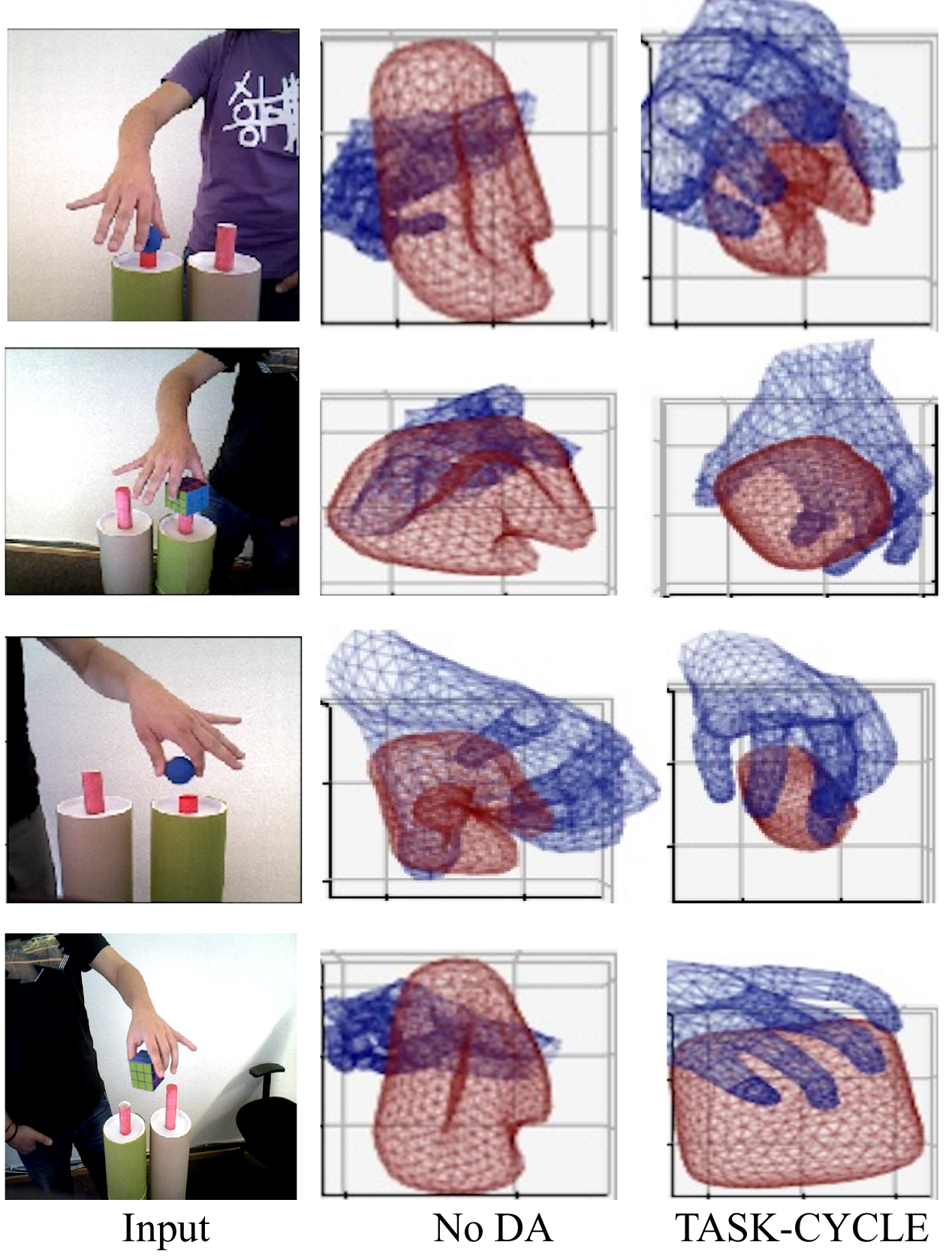}
	\vspace{-2mm}
    \caption{\small {\bf Qualitative results on single images} from the HIC dataset.}	
    \label{fig:result_hic}
	\vspace{-1mm}
\end{figure}

\begin{figure}[!t]
	\centering
	\begin{tabular}{cc}
	\includegraphics[width=0.42\columnwidth]{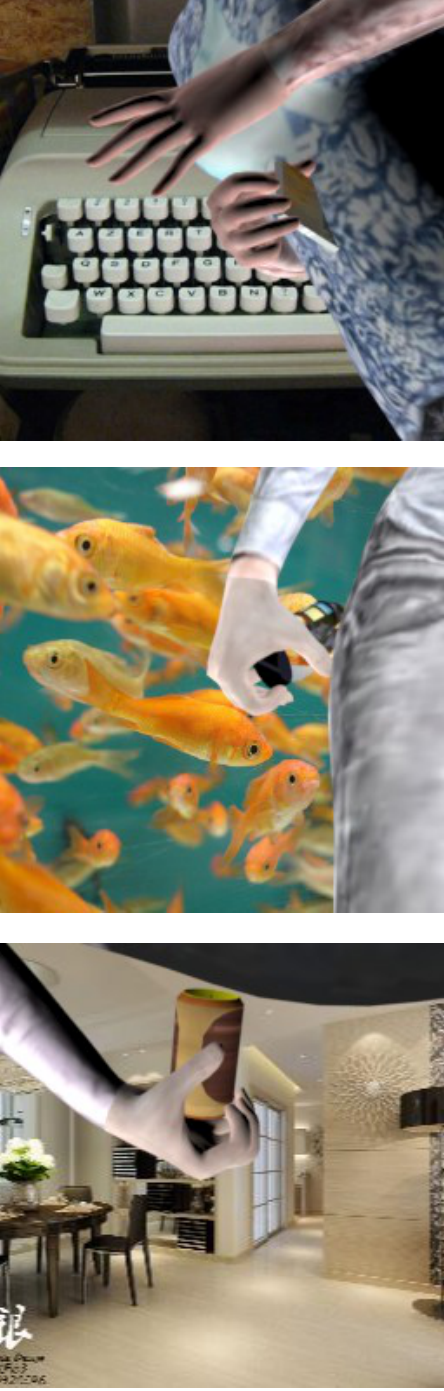}&\includegraphics[width=0.43\columnwidth]{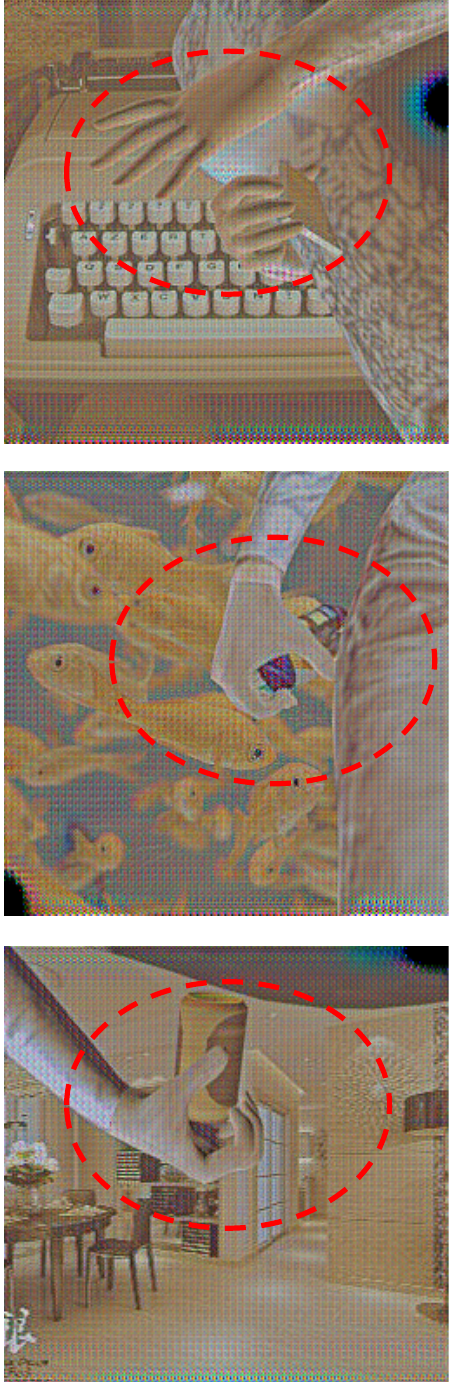}
	\end{tabular}
	\vspace{-1mm}
	\caption{\small {\bf Original and translated ObMan images.} As observed in~\cite{Rozantsev15b}, the translated images do not need to appear particularly realistic to the human eye to be useful for training purposes, as long as their statistics are appropriate for this purpose. }
	\label{fig:translated}
	\vspace{-1mm}
\end{figure}

Fig.~\ref{fig:result_single} depicts four results on individual frames of {\bf FHB}.  In the first three, the recovered hand-pose is very realistic and much better than the baseline {\bf No DA}. The  fourth one showcases a failure. It is attributable to the absence of any such grasping action in the synthetic training set and illustrates the fact that our approach can correct for changing imaging conditions but not for missing poses.   Fig.~\ref{fig:result_temporal} features results over a sequence. Even though each frame is fed to the network individually, the result is very consistent over time. Fig.~\ref{fig:translated} depicts a few of the translated synthetic images created while training the network. Fig.~\ref{fig:result_hic} shows the results on HIC dataset.

\begin{figure*}[t]
	\centering
	\includegraphics[width=0.99\textwidth]{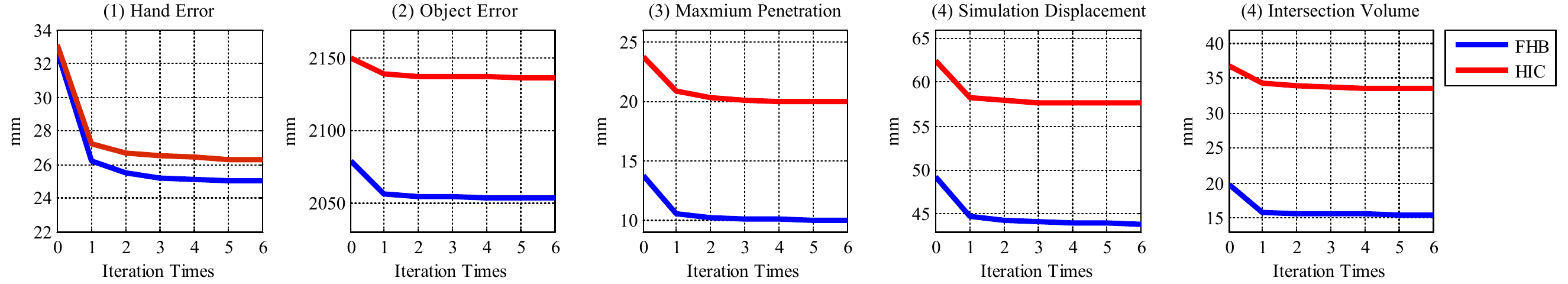}
	\vspace{-2mm}
	\caption{\small {\bf Convergence of our joint training scheme.} Accuracy as a function of the number of iterations for the FHB and HIC datasets.} 
	\label{fig:result_iteration1}
\end{figure*}

We report our quantitative results on {\bf FHB} and {\bf HIC} in Table~\ref{tab:performance_hand_1}. Both \oursC{}, our joint-training approach of Section~\ref{sec:joint}, and  \oursT{}, our full approach with the temporal consistency terms turned on, consistently outperform the baselines, with \oursT{} doing better than \oursC{}. To obtain these results, we performed six iterations of the iterative algorithm of Section~\ref{sec:joint} for each dataset. As can be seen in Fig.~\ref{fig:result_iteration1}, the performance tends to plateau after three iterations. 

To pinpoint the reasons for success or failure, we selected from the FHB test set the 50 images that gave the worst results in terms of hand reconstruction error and the 50 images that gave the best. As can be seen in Table~\ref{tab:performance_hand_2}, in the 50 best cases, our approach divides the error by almost a factor 2, whereas in the 50 worst cases the improvement is much smaller in percentage terms. This closely correlates with the fact that the distance between the observed pose and its closest training sample is an excellent predictor of success or failure. In other words, our approach can compensate for changes of imaging conditions but not for the training set being relatively small and failing to cover the whole range of hand motions that can be seen in the real world.


\begin{table}[tb!]
	\caption{\small {\bf Success vs Failure} on the FHB dataset. {\bf (Top rows)} Hand reconstruction error for the 50 worst and best images, without DA and using \oursC{}.  {\bf (Bottom row)} Distance to the closest pose in the training set. Unsurprisingly, this number correlates closely with the chances of success or failure.}
	\label{tab:performance_hand_2}
	\vspace{-2mm}		
	\centering
		\begin{tabular}{lcc}
			\toprule
			Methods & 50 Worst Images & 50 Best Images \\ \midrule
			{\bf No DA} hand error                                              &59.5          &17.2    \\
			\oursC{}     hand error                                             &56.2         &9.5       \\
			Dist. to closest training sample                                &12.5                     &3.1        \\
			\bottomrule	   
	\end{tabular}
\end{table}

\subsection{Ablation Study}

The bottom rows of Table~\ref{tab:performance_hand_1} document the performance of the variants of our approach introduced in Section~\ref{sec:baseline} and show that the various losses we introduced in Sections~\ref{sec:joint} and~\ref{sec:tempo} all contribute to the final result. In particular, \oursC{} does better than \oursCA{}, which highlights the importance of imposing the task loss at the right place. Also,
\oursT{} does slightly better than \oursTLRender{}, even though we use a much simpler formulation of the long-term temporal loss function. We attribute this to the fact our version is closer to being convex and, unlike the NeuralRenderer, does not introduce any artifacts.

\section{Conclusion}
\label{sec:con}
					
In this paper, we have presented an unsupervised domain adaptation strategy for 3D hand-object joint reconstruction. It involves introducing 3D geometric constraints and self-supervised temporal consistency in a CycleGAN-based framework. Our geometric constraints allow our approach to effectively transfer the annotations from the synthetic source data to the unlabeled, real target domain. Furthermore, our short-term and long-term temporal consistency loss functions let us leverage unlabeled video data to fine-tune the task model. Our experiments on three widely-used datasets have demonstrated the effectiveness of our method and its superiority over the state of the art. In future work, we will explore the use of physics-based constraints to model the contact between hand and object and to supply further supervisory signals.

\ifCLASSOPTIONcompsoc
\section*{Acknowledgments}
\else
\section*{Acknowledgment}
\fi

This work was supported by the Microsoft JRC Project.

			\bibliographystyle{IEEEtran}
			\bibliography{string,vision,learning}

\begin{IEEEbiography}{Mengshi Qi}
	(S'16-M'19) received the B.S. degree from Beijing University of Posts and Telecommunications, Beijing, China, in 2012, and M.S. and Ph.D. degrees in computer science from Beihang University, Beijing, China, in 2014 and 2019, respectively. He is currently a post-doc researcher in the CVLAB at EPFL. His research interests include machine learning and computer vision, especially scene understanding, 3D reconstruction and multimedia analysis.
\end{IEEEbiography}
\vspace{-10mm}

\begin{IEEEbiography}{Edoardo Remelli}
	Currently he is a Ph.D. student and pursuing his Ph.D. degree in Computer Vision Laboratory, EPFL. His research interests lie in the field of computer vision and machine learning with applications in 3D reconstruction.		
\end{IEEEbiography}
\vspace{-10mm}

\begin{IEEEbiography}{Mathieu Salamann} 
	is a Senior Researcher at EPFL. Previously, he was a Senior Researcher and Research Leader in NICTA’s computer vision research group, a Research Assistant Professor at TTI-Chicago, and a postdoctoral fellow at ICSI and EECS at UC Berkeley. He obtained his PhD in Jan. 2009 from EPFL. His research interests lie at the intersection of machine learning and geometry for computer vision.
\end{IEEEbiography}
\vspace{-10mm}

\begin{IEEEbiography}{Pascal Fua}
	is a Professor of Computer Science at EPFL, Switzerland. His research interests include shape and motion reconstruction from images, analysis of microscopy images, and Augmented Reality. He is an IEEE Fellow and has been an Associate Editor of the IEEE journal Transactions for Pattern Analysis and Machine Intelligence.
\end{IEEEbiography}
\vspace{-10mm}

\end{document}